\documentclass{article}
\usepackage{spconf,amsmath,graphicx,hyperref}
\usepackage{booktabs}
\usepackage{multirow}
\usepackage{amsfonts}
\usepackage{colortbl}
\usepackage{xcolor}
\usepackage[normalem]{ulem}

\usepackage{placeins}
\title{Global Time-Aware Conditioning (GTC): A Mechanism for Time-Aware Pediatric Sleep Representation Learning}
\name{Author(s) Name(s)\thanks{Thanks to XYZ agency for funding.}}
\address{Author Affiliation(s)}

\twoauthors
  {Saurav Raj Pandey}
  {Department of Computer Science\\
   University of North Carolina at Chapel Hill\\
   Chapel Hill, NC, USA}
  {Harlin Lee}
  {School of Data and Information Sciences\\
   University of North Carolina at Chapel Hill\\
   Chapel Hill, NC, USA}
\begin{document}
%
\maketitle

\begin{abstract}
Sleep physiology evolves throughout the night, with changes in sleep cycles, respiratory stability, and cardiovascular dynamics. However, self-supervised learning (SSL) methods for sleep typically process short, fixed-duration windows without explicitly modeling where each window occurs within the full sleep session. We introduce Global Time-Aware Conditioning (GTC), a mechanism that complements local positional encodings by conditioning latent representations on the global time-of-sleep of each input window. We integrate GTC into three representative SSL frameworks and evaluate them on five clinically relevant tasks with extreme class imbalance across two independent pediatric cohorts. GTC generally improves F1 and AUPRC across architectures and tasks, with gains persisting under cross-cohort transfer, highlighting the value of global temporal context for sleep representation learning.
\end{abstract}

\section{Introduction}
Sleep is critical to childhood development, with disrupted sleep associated with adverse behavioral, cognitive, and cardiovascular outcomes \cite{sleep_disruption}. Clinical sleep assessment commonly relies on polysomnography (PSG), which records multiple physiologoical signals throughout the night and requires time-consuming expert annotation. 
Self-supervised learning (SSL) provides a promising framework for learning general-purpose representations from unlabeled PSG recordings using pretext objectives such as masked signal reconstruction. These representations can then be transferred to downstream tasks such as sleep stage classification with substantially reduced annotation requirements. 

A key limitation of existing SSL formulations applied to sleep modeling is their reliance on short, fixed-duration windows, such as 30-second epochs, without explicitly accounting for physiological changes across a full night. Respiratory stability, autonomic tone, and cardiac dynamics vary across sleep cycles and time of night \cite{choudhary2009sleep,stein2012heart}. 
Notably, recent work shows that augmenting sleep embeddings with global temporal context after pretraining improves clinical task performance \cite{ye2025uncovering}, motivating the investigation of whether incorporating such context directly during SSL pretraining yields richer representations.


\begin{figure}[t]
    \centering
    \includegraphics[width=\columnwidth]{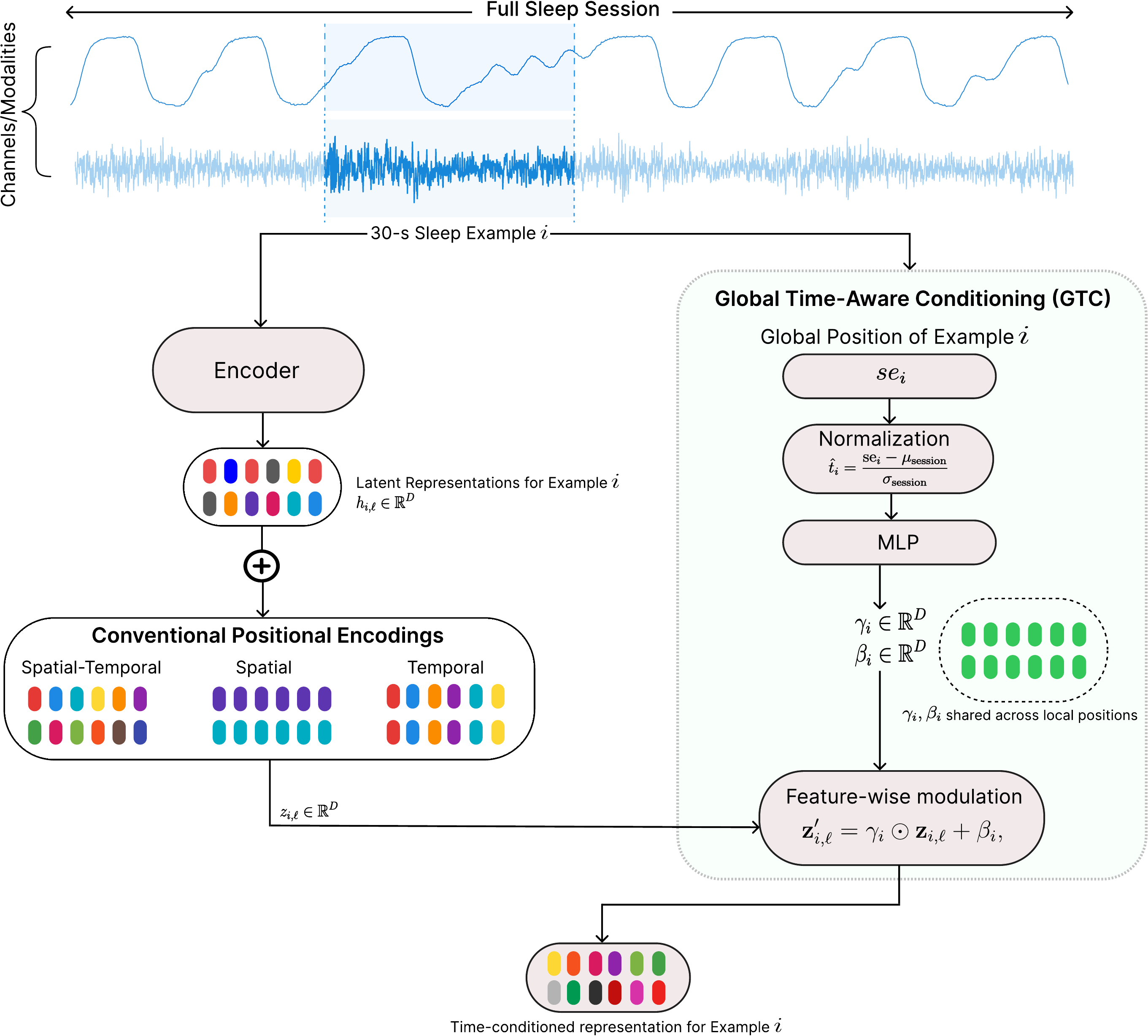}
    \caption{Overview of Global Time-Aware Conditioning (GTC). GTC is agnostic to the number of channels/modalities, the architecture of encoder/backbone, and the existence of decoder.}
    \label{fig:gtc}
\end{figure}

We propose \textbf{Global Time-Aware Conditioning (GTC)}, a lightweight mechanism that complements conventional positional encodings by conditioning latent representations on the global time-of-sleep of each input window, without replacing or modifying existing local encodings. GTC is architecture- and paradigm-agnostic, integrating naturally into diverse SSL frameworks. We integrate GTC into three SSL frameworks spanning distinct backbone architectures and training objectives, and evaluate it across five clinically relevant tasks under severe class imbalance and two independent pediatric cohorts, including cross-cohort transfer from NCH to CHAT.




    

\section{Related Work}
\label{sec:related_work}
Sleep SSL methods have largely focused on unimodal EEG and sleep staging, including BENDR \cite{kostas2021bendr}, EEGPT \cite{wang2024eegpt}, and NeuroNet \cite{lee2024neuronet}, while recent methods such as COCOA \cite{deldari2022cocoa}, SleepFM \cite{sleepfm_original}, SynthSleepNet \cite{synthsleepnet}, and PedSleepMAE \cite{pandey2024pedsleepmae} extend SSL to multimodal physiological signals. These approaches broadly use contrastive learning, which model relationships across views or modalities, or masked prediction, which reconstruct masked inputs from visible observations.


GTC primarily differs from prior SSL methods in how temporal context is incorporated during pretraining. Approaches such as PedSleepMAE, NeuroNet, and SynthSleepNet use standard positional encodings to represent local positions within their input representations, but do not explicitly encode where an input window occurs within its full sleep session.
SleepFM uses an LSTM for sequential modeling during downstream classification, but not during SSL pretraining. While combinations of spatial, temporal, and token encodings altogether have shown promising results on EEG datasets, STAMP \cite{shook2025stamp} and other approaches \cite{irani2502positional} also exclude the tokens' global temporal context. In contrast, GTC complements existing positional encodings by conditioning on the global position of each input window within its sleep session. 



\definecolor{summaryrow}{RGB}{220, 230, 242}
\definecolor{beatgreen}{RGB}{198, 232, 210}

\newcommand{\secondbest}[1]{%
  \begingroup
  \setbox0=\hbox{#1}%
  \rlap{\raisebox{-1.2pt}[0pt][0pt]{\rule{\wd0}{0.35pt}}}%
  \box0\endgroup}
\newcommand{\upcell}[1]{\raisebox{0.7ex}{#1}}
\newcommand{\gpos}[1]{\colorbox{beatgreen}{\strut$+$#1}}
\newcommand{\gnil}[1]{\strut#1}

\begin{table*}[!t]
\centering
\caption{Linear probing on \textbf{NCH} across time-conditioning strategies.
Means $\pm$ SD over seeds 41--43; bold = best, underlined = second-best per metric.
Prevalence shown in parentheses for binary classification tasks.
\colorbox{summaryrow}{\strut Shaded}: $\Delta$F1\,/\,$\Delta$AUPRC of best GTC vs.\ no GTC;
\colorbox{beatgreen}{\strut green} = positive gain.}
\label{tab:nch_results}

\fontsize{9}{10.5}\selectfont
\setlength{\tabcolsep}{1.1pt}
\renewcommand{\arraystretch}{1.0}

\begin{tabular}{
@{}l|l|l|l|
cccc|cccc|cccc|cccc|cccc@{}
}
\hline

\multirow{3}{*}{\textbf{Model}} &
\multirow{3}{*}{\textbf{Arch.}} &
\multirow{3}{*}{\textbf{Train}} &
\multirow{3}{*}{\textbf{GTC}} &
\multicolumn{20}{c}{\textbf{Performance (\%)}} \\

\cline{5-24}

& & & &
\multicolumn{4}{c|}{\rule{0pt}{2ex}\textbf{Sleep Scoring}} &
\multicolumn{4}{c|}{\rule{0pt}{2ex}\textbf{Apnea}} &
\multicolumn{4}{c|}{\rule{0pt}{2ex}\textbf{EEG Arousal}} &
\multicolumn{4}{c|}{\rule{0pt}{2ex}\textbf{Oxygen Desat.}} &
\multicolumn{4}{c}{\rule{0pt}{2ex}\textbf{Hypopnea}} \\

& & & &
\multicolumn{4}{c|}{} &
\multicolumn{4}{c|}{(0.8\%)} &
\multicolumn{4}{c|}{(4.6\%)} &
\multicolumn{4}{c|}{(8.7\%)} &
\multicolumn{4}{c}{(1.9\%)} \\

\cline{5-24}

& & & &
Acc. & F1 & ROC & PR &
Acc. & F1 & ROC & PR &
Acc. & F1 & ROC & PR &
Acc. & F1 & ROC & PR &
Acc. & F1 & ROC & PR \\

\hline

\multirow{4}{*}{SleepFM} &
\multirow{4}{*}{CNN} &
\multirow{4}{*}{CL} &
\rule{0pt}{4.2ex}None &

\shortstack{71.0\\(0.2)} & \shortstack{64.2\\(0.1)} & \shortstack{90.8\\(0.0)} & \shortstack{69.0\\(0.1)} &
\shortstack{97.0\\(0.2)} & \shortstack{17.5\\(0.5)} & \shortstack{\textbf{87.6}\\(0.4)} & \shortstack{11.2\\(0.4)} &
\shortstack{\secondbest{93.8}\\(0.1)} & \shortstack{46.0\\(0.2)} & \shortstack{89.4\\(0.0)} & \shortstack{41.4\\(0.1)} &
\shortstack{\secondbest{78.9}\\(1.3)} & \shortstack{\textbf{32.8}\\(0.5)} & \shortstack{\textbf{77.0}\\(0.2)} & \shortstack{\textbf{25.5}\\(0.2)} &
\shortstack{\secondbest{94.9}\\(0.1)} & \shortstack{20.2\\(0.5)} & \shortstack{\textbf{83.1}\\(0.4)} & \shortstack{12.7\\(0.1)}
\\

\cline{4-24}

& & &
\rule{0pt}{4.2ex}One &

\shortstack{\textbf{71.5}\\(0.1)} & \shortstack{\textbf{64.6}\\(0.0)} & \shortstack{\textbf{91.1}\\(0.0)} & \shortstack{\textbf{69.7}\\(0.1)} &
\shortstack{\secondbest{97.2}\\(0.3)} & \shortstack{\secondbest{18.1}\\(0.5)} & \shortstack{87.4\\(0.1)} & \shortstack{\secondbest{11.4}\\(0.1)} &
\shortstack{93.6\\(0.1)} & \shortstack{\textbf{46.2}\\(0.1)} & \shortstack{\textbf{89.5}\\(0.1)} & \shortstack{\secondbest{41.6}\\(0.3)} &
\shortstack{\secondbest{78.9}\\(1.1)} & \shortstack{32.0\\(0.4)} & \shortstack{76.4\\(0.4)} & \shortstack{24.4\\(0.5)} &
\shortstack{94.5\\(1.1)} & \shortstack{\secondbest{21.0}\\(0.3)} & \shortstack{\secondbest{82.6}\\(0.4)} & \shortstack{\textbf{13.4}\\(0.4)}
\\

\cline{4-24}

& & &
\rule{0pt}{4.2ex}All &

\shortstack{\secondbest{71.4}\\(0.1)} & \shortstack{\secondbest{64.4}\\(0.0)} & \shortstack{\secondbest{91.1}\\(0.0)} & \shortstack{\secondbest{69.6}\\(0.1)} &
\shortstack{\textbf{97.3}\\(0.5)} & \shortstack{\textbf{18.4}\\(0.9)} & \shortstack{\secondbest{87.5}\\(0.2)} & \shortstack{\textbf{11.5}\\(0.3)} &
\shortstack{\textbf{93.9}\\(0.2)} & \shortstack{\secondbest{46.1}\\(0.2)} & \shortstack{\secondbest{89.5}\\(0.1)} & \shortstack{\textbf{41.9}\\(0.2)} &
\shortstack{\textbf{79.4}\\(1.2)} & \shortstack{\secondbest{32.3}\\(0.5)} & \shortstack{\secondbest{76.5}\\(0.4)} & \shortstack{\secondbest{24.8}\\(0.5)} &
\shortstack{\textbf{94.9}\\(0.4)} & \shortstack{\textbf{21.2}\\(0.4)} & \shortstack{\secondbest{82.6}\\(0.4)} & \shortstack{\secondbest{13.3}\\(0.3)}
\\

\cline{4-24}

& & &
\multicolumn{1}{l|}{\cellcolor{summaryrow}\footnotesize $\Delta$F1\,/\,$\Delta$PR} &
\multicolumn{4}{c|}{\cellcolor{summaryrow}\gpos{0.4}\,/\,\gpos{0.7}} &
\multicolumn{4}{c|}{\cellcolor{summaryrow}\gpos{0.9}\,/\,\gpos{0.3}} &
\multicolumn{4}{c|}{\cellcolor{summaryrow}\gpos{0.2}\,/\,\gpos{0.5}} &
\multicolumn{4}{c|}{\cellcolor{summaryrow}\gnil{$-$0.5}\,/\,\gnil{$-$0.7}} &
\multicolumn{4}{c}{\cellcolor{summaryrow}\gpos{1.0}\,/\,\gpos{0.7}}
\\

\hline

\multirow{4}{*}{\shortstack{PedSleep-\\MAE}} &
\multirow{4}{*}{ViT} &
\multirow{4}{*}{MP} &
\rule{0pt}{4.2ex}None &

\shortstack{70.6\\(0.3)} & \shortstack{61.8\\(0.1)} & \shortstack{91.0\\(0.1)} & \shortstack{65.8\\(0.1)} &
\shortstack{\secondbest{97.0}\\(0.2)} & \shortstack{\secondbest{13.9}\\(0.2)} & \shortstack{\secondbest{81.6}\\(0.0)} & \shortstack{\secondbest{6.3}\\(0.1)} &
\shortstack{\textbf{90.1}\\(0.3)} & \shortstack{\textbf{29.0}\\(0.1)} & \shortstack{\textbf{78.1}\\(0.1)} & \shortstack{\textbf{19.4}\\(0.0)} &
\shortstack{\secondbest{82.1}\\(0.4)} & \shortstack{32.4\\(0.0)} & \shortstack{76.3\\(0.0)} & \shortstack{25.5\\(0.0)} &
\shortstack{\secondbest{93.6}\\(0.2)} & \shortstack{18.6\\(0.1)} & \shortstack{\secondbest{80.2}\\(0.2)} & \shortstack{10.4\\(0.1)}
\\

\cline{4-24}

& & &
\rule{0pt}{4.2ex}Enc. &

\shortstack{\textbf{72.7}\\(0.1)} & \shortstack{\textbf{64.1}\\(0.0)} & \shortstack{\textbf{92.0}\\(0.0)} & \shortstack{\textbf{68.1}\\(0.1)} &
\shortstack{96.4\\(0.0)} & \shortstack{12.7\\(0.1)} & \shortstack{81.5\\(0.2)} & \shortstack{6.0\\(0.1)} &
\shortstack{\secondbest{88.7}\\(0.1)} & \shortstack{\secondbest{27.5}\\(0.1)} & \shortstack{\secondbest{77.3}\\(0.0)} & \shortstack{16.9\\(0.1)} &
\shortstack{81.5\\(0.1)} & \shortstack{\secondbest{33.4}\\(0.0)} & \shortstack{\textbf{77.2}\\(0.1)} & \shortstack{\textbf{27.0}\\(0.1)} &
\shortstack{93.4\\(0.2)} & \shortstack{\textbf{18.9}\\(0.1)} & \shortstack{\textbf{80.6}\\(0.0)} & \shortstack{\textbf{10.9}\\(0.1)}
\\

\cline{4-24}

& & &
\shortstack{Enc.\\\&\\Dec.} &

\upcell{\shortstack{\secondbest{72.4}\\(0.3)}} & \upcell{\shortstack{\secondbest{63.5}\\(0.1)}} & \upcell{\shortstack{\secondbest{91.7}\\(0.0)}} & \upcell{\shortstack{\secondbest{67.5}\\(0.1)}} &
\upcell{\shortstack{\textbf{97.9}\\(0.1)}} & \upcell{\shortstack{\textbf{14.6}\\(0.1)}} & \upcell{\shortstack{\textbf{82.3}\\(0.0)}} & \upcell{\shortstack{\textbf{6.5}\\(0.0)}} &
\upcell{\shortstack{87.9\\(0.2)}} & \upcell{\shortstack{26.9\\(0.1)}} & \upcell{\shortstack{77.2\\(0.0)}} & \upcell{\shortstack{\secondbest{17.7}\\(0.0)}} &
\upcell{\shortstack{\textbf{83.2}\\(0.2)}} & \upcell{\shortstack{\textbf{33.4}\\(0.1)}} & \upcell{\shortstack{\secondbest{77.2}\\(0.0)}} & \upcell{\shortstack{\secondbest{26.7}\\(0.1)}} &
\upcell{\shortstack{\textbf{94.9}\\(0.2)}} & \upcell{\shortstack{\secondbest{18.8}\\(0.1)}} & \upcell{\shortstack{80.1\\(0.1)}} & \upcell{\shortstack{\secondbest{10.6}\\(0.0)}}
\\

\cline{4-24}

& & &
\multicolumn{1}{l|}{\cellcolor{summaryrow}\footnotesize $\Delta$F1\,/\,$\Delta$PR} &
\multicolumn{4}{c|}{\cellcolor{summaryrow}\gpos{2.3}\,/\,\gpos{2.3}} &
\multicolumn{4}{c|}{\cellcolor{summaryrow}\gpos{0.7}\,/\,\gpos{0.2}} &
\multicolumn{4}{c|}{\cellcolor{summaryrow}\gnil{$-$1.5}\,/\,\gnil{$-$1.7}} &
\multicolumn{4}{c|}{\cellcolor{summaryrow}\gpos{1.0}\,/\,\gpos{1.5}} &
\multicolumn{4}{c}{\cellcolor{summaryrow}\gpos{0.3}\,/\,\gpos{0.5}}
\\

\hline

\multirow{4}{*}{\shortstack{Synth-\\SleepNet}} &
\multirow{4}{*}{\shortstack{Neuro-\\Net}} &
\multirow{4}{*}{\shortstack{CL\\+\\MP}} &
\rule{0pt}{4.2ex}None &

\shortstack{\textbf{75.3}\\(0.2)} & \shortstack{\textbf{68.3}\\(0.2)} & \shortstack{\textbf{93.3}\\(0.1)} & \shortstack{\textbf{73.9}\\(0.1)} &
\shortstack{\textbf{94.3}\\(1.5)} & \shortstack{\secondbest{7.2}\\(0.3)} & \shortstack{\secondbest{77.1}\\(1.0)} & \shortstack{\secondbest{3.3}\\(0.2)} &
\shortstack{\textbf{93.7}\\(0.2)} & \shortstack{\textbf{45.2}\\(0.0)} & \shortstack{\secondbest{89.1}\\(0.1)} & \shortstack{\textbf{39.9}\\(0.0)} &
\shortstack{69.5\\(0.4)} & \shortstack{26.5\\(0.1)} & \shortstack{71.0\\(0.1)} & \shortstack{17.6\\(0.2)} &
\shortstack{\textbf{90.9}\\(0.7)} & \shortstack{12.3\\(0.2)} & \shortstack{\secondbest{75.6}\\(0.5)} & \shortstack{\secondbest{6.3}\\(0.2)}
\\

\cline{4-24}

& & &
\rule{0pt}{4.2ex}Enc. &

\shortstack{\textbf{75.3}\\(0.2)} & \shortstack{\textbf{68.3}\\(0.2)} & \shortstack{\textbf{93.3}\\(0.1)} & \shortstack{\secondbest{73.7}\\(0.1)} &
\shortstack{\textbf{94.3}\\(0.9)} & \shortstack{7.1\\(0.2)} & \shortstack{76.9\\(0.1)} & \shortstack{3.2\\(0.0)} &
\shortstack{\textbf{93.7}\\(0.0)} & \shortstack{\secondbest{45.1}\\(0.1)} & \shortstack{\textbf{89.3}\\(0.0)} & \shortstack{\secondbest{39.8}\\(0.1)} &
\shortstack{\secondbest{70.7}\\(0.8)} & \shortstack{\secondbest{27.0}\\(0.3)} & \shortstack{\secondbest{71.6}\\(0.3)} & \shortstack{\textbf{18.8}\\(0.4)} &
\shortstack{\secondbest{90.0}\\(1.2)} & \shortstack{\textbf{12.7}\\(0.4)} & \shortstack{\textbf{76.7}\\(0.4)} & \shortstack{\textbf{6.8}\\(0.2)}
\\

\cline{4-24}

& & &
\shortstack{Enc.\\\&\\Dec.} &

\upcell{\shortstack{\secondbest{75.1}\\(0.3)}} & \upcell{\shortstack{\secondbest{68.1}\\(0.2)}} & \upcell{\shortstack{\secondbest{93.1}\\(0.0)}} & \upcell{\shortstack{73.5\\(0.0)}} &
\upcell{\shortstack{\secondbest{93.7}\\(0.4)}} & \upcell{\shortstack{\textbf{8.0}\\(0.3)}} & \upcell{\shortstack{\textbf{78.1}\\(0.3)}} & \upcell{\shortstack{\textbf{4.0}\\(0.2)}} &
\upcell{\shortstack{\secondbest{93.3}\\(0.1)}} & \upcell{\shortstack{42.7\\(0.1)}} & \upcell{\shortstack{88.0\\(0.0)}} & \upcell{\shortstack{37.0\\(0.1)}} &
\upcell{\shortstack{\textbf{71.2}\\(1.2)}} & \upcell{\shortstack{\textbf{27.1}\\(0.2)}} & \upcell{\shortstack{\textbf{71.7}\\(0.3)}} & \upcell{\shortstack{\secondbest{18.5}\\(0.3)}} &
\upcell{\shortstack{87.8\\(0.4)}} & \upcell{\shortstack{\secondbest{12.4}\\(0.2)}} & \upcell{\shortstack{\textbf{76.7}\\(0.2)}} & \upcell{\shortstack{\textbf{6.8}\\(0.2)}}
\\

\cline{4-24}

& & &
\multicolumn{1}{l|}{\cellcolor{summaryrow}\footnotesize $\Delta$F1\,/\,$\Delta$PR} &
\multicolumn{4}{c|}{\cellcolor{summaryrow}\gnil{$0.0$}\,/\,\gnil{$-$0.2}} &
\multicolumn{4}{c|}{\cellcolor{summaryrow}\gpos{0.8}\,/\,\gpos{0.7}} &
\multicolumn{4}{c|}{\cellcolor{summaryrow}\gnil{$-$0.1}\,/\,\gnil{$-$0.1}} &
\multicolumn{4}{c|}{\cellcolor{summaryrow}\gpos{0.6}\,/\,\gpos{1.2}} &
\multicolumn{4}{c}{\cellcolor{summaryrow}\gpos{0.4}\,/\,\gpos{0.5}}
\\

\hline

\end{tabular}

\smallskip

\parbox{\textwidth}{
\fontsize{9}{10.5}\selectfont
\textit{Note:}
Arch.: backbone architecture. Train: CL = contrastive learning, MP = masked prediction. GTC: placement of global time conditioning. For SleepFM, ``One'' conditions RESP for respiratory tasks and BAS for sleep staging/arousal. F1: macro-F1 (sleep staging), binary F1 (events). ROC/PR: AUROC/AUPRC.
}

\end{table*}

\begin{table*}[htbp]
\centering
\caption{Linear probing on \textbf{CHAT} across time-conditioning strategies.
Layout and conventions as in Table~\ref{tab:nch_results}.}
\label{tab:chat-final-test}

\fontsize{9}{10.5}\selectfont
\setlength{\tabcolsep}{1.1pt}
\renewcommand{\arraystretch}{1.0}

\begin{tabular}{
@{}l|l|
cccc|cccc|cccc|cccc|cccc@{}
}
\hline

\multirow{3}{*}{\textbf{Model}} &
\multirow{3}{*}{\textbf{GTC}} &
\multicolumn{20}{c}{\textbf{Performance (\%)}} \\

\cline{3-22}

& &
\multicolumn{4}{c|}{\rule{0pt}{2ex}\textbf{Sleep Scoring}} &
\multicolumn{4}{c|}{\rule{0pt}{2ex}\textbf{Apnea}} &
\multicolumn{4}{c|}{\rule{0pt}{2ex}\textbf{EEG Arousal}} &
\multicolumn{4}{c|}{\rule{0pt}{2ex}\textbf{Oxygen Desat.}} &
\multicolumn{4}{c}{\rule{0pt}{2ex}\textbf{Hypopnea}} \\

& &
\multicolumn{4}{c|}{} &
\multicolumn{4}{c|}{(2.4\%)} &
\multicolumn{4}{c|}{(8.0\%)} &
\multicolumn{4}{c|}{(17.0\%)} &
\multicolumn{4}{c}{(3.9\%)} \\

\cline{3-22}

& &
Acc. & F1 & ROC & PR &
Acc. & F1 & ROC & PR &
Acc. & F1 & ROC & PR &
Acc. & F1 & ROC & PR &
Acc. & F1 & ROC & PR \\

\hline

\multirow{4}{*}{SleepFM} &
\rule{0pt}{4.2ex}None &

\shortstack{76.6\\(0.2)} & \shortstack{71.2\\(0.2)} & \shortstack{94.0\\(0.0)} & \shortstack{76.3\\(0.0)} &
\shortstack{\textbf{94.2}\\(0.2)} & \shortstack{\secondbest{14.7}\\(0.2)} & \shortstack{81.5\\(0.1)} & \shortstack{\textbf{8.6}\\(0.1)} &
\shortstack{\secondbest{92.3}\\(0.3)} & \shortstack{58.1\\(0.1)} & \shortstack{92.9\\(0.1)} & \shortstack{57.2\\(0.3)} &
\shortstack{76.4\\(0.6)} & \shortstack{\secondbest{44.6}\\(0.3)} & \shortstack{\textbf{77.7}\\(0.3)} & \shortstack{41.1\\(0.7)} &
\shortstack{\textbf{91.9}\\(0.8)} & \shortstack{\secondbest{21.4}\\(0.5)} & \shortstack{80.1\\(0.4)} & \shortstack{13.8\\(0.4)}
\\

\cline{2-22}

&
\rule{0pt}{4.2ex}One &

\shortstack{\secondbest{78.0}\\(0.3)} & \shortstack{\secondbest{72.8}\\(0.3)} & \shortstack{\secondbest{94.8}\\(0.1)} & \shortstack{\secondbest{78.5}\\(0.1)} &
\shortstack{91.8\\(0.4)} & \shortstack{14.6\\(0.3)} & \shortstack{\textbf{81.7}\\(0.4)} & \shortstack{\secondbest{8.3}\\(0.1)} &
\shortstack{\textbf{93.0}\\(0.2)} & \shortstack{\secondbest{61.6}\\(0.0)} & \shortstack{\secondbest{93.9}\\(0.0)} & \shortstack{\secondbest{62.1}\\(0.2)} &
\shortstack{\textbf{77.3}\\(0.1)} & \shortstack{\secondbest{44.6}\\(0.4)} & \shortstack{77.4\\(0.3)} & \shortstack{\secondbest{42.0}\\(0.6)} &
\shortstack{\secondbest{90.4}\\(0.2)} & \shortstack{\textbf{21.6}\\(0.1)} & \shortstack{\textbf{81.5}\\(0.3)} & \shortstack{\textbf{14.3}\\(0.3)}
\\

\cline{2-22}

&
\rule{0pt}{4.2ex}All &

\shortstack{\textbf{78.1}\\(0.2)} & \shortstack{\textbf{73.0}\\(0.2)} & \shortstack{\textbf{94.9}\\(0.1)} & \shortstack{\textbf{78.8}\\(0.1)} &
\shortstack{\secondbest{92.0}\\(0.9)} & \shortstack{\textbf{14.8}\\(0.1)} & \shortstack{\secondbest{81.6}\\(0.4)} & \shortstack{\secondbest{8.3}\\(0.1)} &
\shortstack{\textbf{93.0}\\(0.3)} & \shortstack{\textbf{62.0}\\(0.2)} & \shortstack{\textbf{94.0}\\(0.0)} & \shortstack{\textbf{62.8}\\(0.1)} &
\shortstack{\secondbest{76.7}\\(1.2)} & \shortstack{\textbf{44.8}\\(0.3)} & \shortstack{\secondbest{77.5}\\(0.1)} & \shortstack{\textbf{42.3}\\(0.1)} &
\shortstack{90.3\\(0.3)} & \shortstack{\textbf{21.6}\\(0.2)} & \shortstack{\secondbest{81.4}\\(0.3)} & \shortstack{\secondbest{14.1}\\(0.3)}
\\

\cline{2-22}

&
\multicolumn{1}{l|}{\cellcolor{summaryrow}\footnotesize $\Delta$F1\,/\,$\Delta$PR} &
\multicolumn{4}{c|}{\cellcolor{summaryrow}\gpos{1.8}\,/\,\gpos{2.5}} &
\multicolumn{4}{c|}{\cellcolor{summaryrow}\gpos{0.1}\,/\,\gnil{$-$0.3}} &
\multicolumn{4}{c|}{\cellcolor{summaryrow}\gpos{3.9}\,/\,\gpos{5.6}} &
\multicolumn{4}{c|}{\cellcolor{summaryrow}\gpos{0.2}\,/\,\gpos{1.2}} &
\multicolumn{4}{c}{\cellcolor{summaryrow}\gpos{0.2}\,/\,\gpos{0.5}}
\\

\hline

\multirow{4}{*}{\shortstack{PedSleepMAE}} &
\rule{0pt}{4.2ex}None &

\shortstack{39.8\\(4.3)} & \shortstack{22.8\\(1.4)} & \shortstack{61.8\\(0.1)} & \shortstack{29.7\\(0.1)} &
\shortstack{11.7\\(0.0)} & \shortstack{4.5\\(0.0)} & \shortstack{43.9\\(0.0)} & \shortstack{1.9\\(0.0)} &
\shortstack{\textbf{87.8}\\(0.2)} & \shortstack{35.3\\(0.3)} & \shortstack{69.3\\(0.0)} & \shortstack{21.8\\(0.0)} &
\shortstack{26.1\\(0.0)} & \shortstack{30.7\\(0.0)} & \shortstack{40.4\\(0.0)} & \shortstack{13.4\\(0.0)} &
\shortstack{\textbf{93.8}\\(1.0)} & \shortstack{\textbf{18.8}\\(0.7)} & \shortstack{55.7\\(0.5)} & \shortstack{\textbf{11.0}\\(1.3)}
\\

\cline{2-22}

&
\rule{0pt}{4.2ex}Enc. &

\shortstack{\textbf{45.2}\\(4.5)} & \shortstack{\secondbest{37.0}\\(2.0)} & \shortstack{\secondbest{72.3}\\(1.2)} & \shortstack{\secondbest{37.2}\\(1.2)} &
\shortstack{\textbf{86.7}\\(5.2)} & \shortstack{\secondbest{5.9}\\(0.0)} & \shortstack{\secondbest{59.8}\\(0.2)} & \shortstack{\secondbest{3.2}\\(0.0)} &
\shortstack{\secondbest{87.3}\\(0.2)} & \shortstack{\secondbest{35.5}\\(0.2)} & \shortstack{\textbf{73.9}\\(0.1)} & \shortstack{\secondbest{24.1}\\(0.0)} &
\shortstack{\secondbest{47.6}\\(2.0)} & \shortstack{\secondbest{31.5}\\(0.1)} & \shortstack{\secondbest{62.1}\\(0.1)} & \shortstack{\secondbest{23.1}\\(0.1)} &
\shortstack{46.5\\(0.2)} & \shortstack{7.7\\(0.0)} & \shortstack{\secondbest{59.3}\\(0.0)} & \shortstack{4.5\\(0.0)}
\\

\cline{2-22}

&
\rule{0pt}{4.2ex}\shortstack{Enc. \& Dec.} &

{\shortstack{\textbf{45.2}\\(1.3)}} & {\shortstack{\textbf{37.7}\\(1.8)}} & {\shortstack{\textbf{73.6}\\(0.6)}} & {\shortstack{\textbf{39.6}\\(0.6)}} &
{\shortstack{\secondbest{84.3}\\(0.6)}} & {\shortstack{\textbf{6.9}\\(0.1)}} & {\shortstack{\textbf{62.5}\\(0.2)}} & {\shortstack{\textbf{3.3}\\(0.0)}} &
{\shortstack{86.6\\(0.3)}} & {\shortstack{\textbf{35.9}\\(0.1)}} & {\shortstack{\secondbest{73.6}\\(0.3)}} & {\shortstack{\textbf{24.2}\\(0.4)}} &
{\shortstack{\textbf{69.9}\\(1.1)}} & {\shortstack{\textbf{40.2}\\(0.1)}} & {\shortstack{\textbf{72.4}\\(0.1)}} & {\shortstack{\textbf{33.1}\\(0.1)}} &
{\shortstack{\secondbest{88.5}\\(1.0)}} & {\shortstack{\secondbest{15.8}\\(0.5)}} & {\shortstack{\textbf{71.7}\\(0.3)}} & {\shortstack{\secondbest{8.6}\\(0.3)}}
\\

\cline{2-22}

&
\multicolumn{1}{l|}{\cellcolor{summaryrow}\footnotesize $\Delta$F1\,/\,$\Delta$PR} &
\multicolumn{4}{c|}{\cellcolor{summaryrow}\gpos{14.9}\,/\,\gpos{9.9}} &
\multicolumn{4}{c|}{\cellcolor{summaryrow}\gpos{2.4}\,/\,\gpos{1.4}} &
\multicolumn{4}{c|}{\cellcolor{summaryrow}\gpos{0.6}\,/\,\gpos{2.4}} &
\multicolumn{4}{c|}{\cellcolor{summaryrow}\gpos{9.5}\,/\,\gpos{19.7}} &
\multicolumn{4}{c}{\cellcolor{summaryrow}\gnil{$-$3.0}\,/\,\gnil{$-$2.4}}
\\

\hline

\multirow{4}{*}{\shortstack{SynthSleepNet}} &
\rule{0pt}{4.2ex}None &

\shortstack{\textbf{77.5}\\(0.1)} & \shortstack{\textbf{72.6}\\(0.1)} & \shortstack{\secondbest{94.1}\\(0.0)} & \shortstack{\textbf{77.2}\\(0.1)} &
\shortstack{\textbf{93.5}\\(0.5)} & \shortstack{\secondbest{14.2}\\(0.1)} & \shortstack{\textbf{81.4}\\(0.1)} & \shortstack{\textbf{8.3}\\(0.1)} &
\shortstack{\textbf{92.5}\\(0.1)} & \shortstack{\secondbest{58.9}\\(0.2)} & \shortstack{\secondbest{92.6}\\(0.0)} & \shortstack{\secondbest{56.8}\\(0.1)} &
\shortstack{\secondbest{75.6}\\(1.2)} & \shortstack{\secondbest{39.8}\\(0.6)} & \shortstack{\textbf{74.1}\\(0.1)} & \shortstack{\secondbest{33.9}\\(0.1)} &
\shortstack{\textbf{90.3}\\(0.4)} & \shortstack{\textbf{18.8}\\(0.1)} & \shortstack{\textbf{80.0}\\(0.0)} & \shortstack{\textbf{11.4}\\(0.1)}
\\

\cline{2-22}

&
\rule{0pt}{4.2ex}Enc. &

\shortstack{\secondbest{77.3}\\(0.1)} & \shortstack{\secondbest{72.4}\\(0.0)} & \shortstack{\textbf{94.2}\\(0.0)} & \shortstack{76.9\\(0.1)} &
\shortstack{91.6\\(1.0)} & \shortstack{\secondbest{14.2}\\(0.3)} & \shortstack{\secondbest{80.6}\\(0.1)} & \shortstack{\textbf{8.3}\\(0.0)} &
\shortstack{\textbf{92.5}\\(0.1)} & \shortstack{\textbf{59.0}\\(0.1)} & \shortstack{\textbf{92.9}\\(0.0)} & \shortstack{\textbf{58.0}\\(0.1)} &
\shortstack{\secondbest{75.6}\\(0.3)} & \shortstack{\secondbest{39.8}\\(0.2)} & \shortstack{\secondbest{73.6}\\(0.1)} & \shortstack{\textbf{34.1}\\(0.2)} &
\shortstack{89.6\\(0.8)} & \shortstack{\secondbest{18.7}\\(0.0)} & \shortstack{\secondbest{79.4}\\(0.1)} & \shortstack{\textbf{11.4}\\(0.0)}
\\

\cline{2-22}

&
\rule{0pt}{4.2ex}\shortstack{Enc. \& Dec.} &

{\shortstack{\secondbest{77.3}\\(0.1)}} & {\shortstack{72.3\\(0.0)}} & {\shortstack{\secondbest{94.1}\\(0.0)}} & {\shortstack{\secondbest{77.0}\\(0.0)}} &
{\shortstack{\secondbest{92.1}\\(0.6)}} & {\shortstack{\textbf{14.3}\\(0.0)}} & {\shortstack{79.9\\(0.1)}} & {\shortstack{\secondbest{7.8}\\(0.0)}} &
{\shortstack{\secondbest{90.7}\\(0.2)}} & {\shortstack{54.7\\(0.2)}} & {\shortstack{91.6\\(0.0)}} & {\shortstack{51.6\\(0.0)}} &
{\shortstack{\textbf{76.0}\\(0.4)}} & {\shortstack{\textbf{40.3}\\(0.5)}} & {\shortstack{\textbf{74.1}\\(0.1)}} & {\shortstack{33.8\\(0.0)}} &
{\shortstack{\secondbest{89.8}\\(0.4)}} & {\shortstack{18.3\\(0.1)}} & {\shortstack{78.9\\(0.1)}} & {\shortstack{\secondbest{11.0}\\(0.0)}}
\\

\cline{2-22}

&
\multicolumn{1}{l|}{\cellcolor{summaryrow}\footnotesize $\Delta$F1\,/\,$\Delta$PR} &
\multicolumn{4}{c|}{\cellcolor{summaryrow}\gnil{$-$0.2}\,/\,\gnil{$-$0.2}} &
\multicolumn{4}{c|}{\cellcolor{summaryrow}\gpos{0.1}\,/\,\gnil{$0.0$}} &
\multicolumn{4}{c|}{\cellcolor{summaryrow}\gpos{0.1}\,/\,\gpos{1.2}} &
\multicolumn{4}{c|}{\cellcolor{summaryrow}\gpos{0.5}\,/\,\gpos{0.2}} &
\multicolumn{4}{c}{\cellcolor{summaryrow}\gnil{$-$0.1}\,/\,\gnil{$0.0$}}
\\

\hline

\end{tabular}

\end{table*}

\section{Methodology}
We describe the datasets we used and the GTC mechanism. 
All code will be made public upon acceptance of the paper. 

\subsection{Dataset and Preprocessing}
We use Nationwide Children’s Hospital (NCH) Sleep Databank, a large publicly available pediatric PSG dataset collected in real-world clinical settings \cite{lee2022large} to train our models with and without GTC. We analyze 2{,}379 overnight PSG recordings containing a consistent set of sixteen commonly available physiological channels, including EEG, EOG, and respiratory signals. 

The Childhood Adenotonsillectomy Trial (CHAT) is a multi-center randomized controlled study of children with mild to moderate obstructive sleep apnea with standardized overnight PSG and centralized scoring \cite{zhang2018national, marcus2013randomized}. We use this cohort to fine-tune our NCH-trained models on CHAT's baseline subset collected prior to any intervention. To ensure compatibility with NCH, we restrict CHAT to 12 overlapping physiological channels, which comprises 422 PSGs. 

For both datasets, expert annotations are provided at 30-second resolution, including sleep stages (Wake, Non-REM 1, 2, 3, and REM) and clinically relevant events such as apnea, hypopnea, arousal, and oxygen desaturation. Each non-overlapping 30-second epoch is treated as an individual sample for training and inference. We standardize each physiological channel to zero mean and unit standard deviation to facilitate stable model training. We perform a 80-10-10 split for training, validation, and testing at the patient level to avoid data leakage.



\subsection{Global Time-Aware Conditioning (GTC)}
Conventional positional encodings describe the \emph{local} position of representations within an input, but do not indicate where the input itself occurs within the full sleep session. GTC complements them with a sample-level global time signal representing the position of each 30-second window within the recording. This allows otherwise identical windows to be seen differently based on their position within the night. 

We represent global time using the sequential location of a 30-second window in its corresponding sleep recording. For example $i$, we denote this segment index by $\mathrm{se}_i$. Because recording durations vary across patients, we normalize the segment index using the dataset-level mean and standard deviation of session lengths:
\begin{equation}
    \hat{t}_i =
    \frac{\mathrm{se}_i - \mu_{\text{session}}}
    {\sigma_{\text{session}}},
\end{equation}
where $\mu_{\text{session}}$ and $\sigma_{\text{session}}$ are computed from the number of examples in each sleep session.

The normalized global-time value $\hat{t}_i$ is passed through a two-layer MLP to generate feature-wise modulation parameters $(\gamma_i,\beta_i) \in \mathbb{R}^{D}$, where $D$ is the embedding dimension of the latent tokens. The resulting vectors are layer-normalized and controlled by learnable scalar gates initialized to zero, such that introducing GTC initially preserves the unconditioned representation.

Let $\mathbf{z}_{i,\ell} \in \mathbb{R}^{D}$ denote the latent representation at local position $\ell$ within example $i$. After positional encoding, GTC modulates each latent representation as
\begin{equation}
    \mathbf{z}'_{i,\ell}
    =
    \gamma_i \odot \mathbf{z}_{i,\ell} + \beta_i,
    \label{eq:time_film}
\end{equation}
where the same $(\gamma_i,\beta_i)$ pair is applied across all local latent positions within example $i$. This formulation follows feature-wise affine modulation~\cite{perez2018film}, while using global time-of-sleep as the conditioning signal. Thus, GTC complements local positional information with the position of the example within the full sleep session. Furthermore, GTC is architecture-agnostic, requiring only a small number of additional parameters that scale linearly with $D$. Figure \ref{fig:gtc} illustrates this mechanism on a sample input.

\subsection{GTC Integration Across SSL Architectures}
We adapt the placement of GTC to the representation pipeline of each SSL architecture and evaluate with linear probing.

SleepFM uses multimodal contrastive pretraining with a CNN encoder, channel pooling, a temporal Transformer, and temporal pooling. We apply GTC after temporal positional encoding and before the temporal Transformer. The 16 PSG channels are grouped into three modalities: BAS, RESP, and EMG. To prevent global time from serving as a shortcut during contrastive pretraining, only one randomly selected modality receives GTC in each mini-batch. At evaluation, we condition either the task-relevant modality or all three.

PedSleepMAE follows a masked autoencoding framework with Vision Transformer (ViT)-based encoder and decoder networks. In encoder-only, GTC is applied after patch embedding and positional encoding but before masking. In encoder-decoder, GTC is additionally applied to the decoder representations before reconstruction.

SynthSleepNet combines masked autoencoding and contrastive learning, using modality-specific pretrained NeuroNet encoders to obtain unimodal representations \cite{lee2024neuronet}. We evaluate two GTC configurations. In encoder-only, GTC is applied to each unimodal representation after positional encoding and before masking. In encoder-decoder, GTC is additionally applied in the decoder before masked reconstruction. Although SynthSleepNet supports several input configurations, we use one channel from each of three modalities (EEG, EOG, and EMG), balancing multimodal representation learning with computational cost.

For downstream evaluation on NCH, we use a frozen-backbone protocol, fitting a linear classifier on the learned representations to assess changes in representation quality introduced by GTC. For CHAT, we initialize each model with its NCH-pretrained weights and fine-tune it on CHAT using the corresponding self-supervised objective. We then follow the same linear probing protocol as on NCH. Class-weighted losses are used to account for class imbalance. We keep identical hyperparameters while training models with and without GTC and will list all hyperparameters in our repository.

\section{RESULTS AND DISCUSSION}
Table~\ref{tab:nch_results} summarizes downstream performance across five tasks on NCH. Given the substantial class imbalance and low prevalence of several events, we focus on F1 and AUPRC as more informative measures of performance than accuracy and AUROC. We therefore emphasize improvements relative to the corresponding unconditioned baseline. GTC improves F1 over the unconditioned baseline on four of five tasks for both SleepFM and PedSleepMAE. The largest gains are observed for PedSleepMAE, particularly in sleep scoring, where encoder conditioning improves both F1 and AUPRC by 2.3 pp. GTC also improves PedSleepMAE on apnea, desaturation, and hypopnea, while the unconditioned model performs best for EEG arousal. SynthSleepNet shows smaller improvements, with F1 gains on apnea, desaturation, and hypopnea. This may reflect its combined CL and MP objective at the multimodal level, as well as at its pretrained unimodal NeuroNet encoders, which together produce richer representations that leave less room for GTC to improve.

Table~\ref{tab:chat-final-test} evaluates cross-cohort generalization by fine-tuning NCH-pretrained models on CHAT. SleepFM shows consistent F1 gains across all five tasks, most notably for sleep scoring (+1.8 pp) and EEG arousal (+3.9 pp). PedSleepMAE shows the largest gains, improving F1 on four tasks, including nearly 15 pp for sleep scoring and 10 pp for oxygen desaturation. Notably, GTC gains for PedSleepMAE are substantially larger on CHAT than on NCH, suggesting that global temporal structure learned during pretraining may be particularly beneficial when adapting to a new cohort. These improvements suggest that representations learned with GTC transfer effectively across cohorts under full fine-tuning. SynthSleepNet shows smaller gains, improving F1 on three of five tasks.


While the optimal GTC configuration is task- and architecture-dependent, our results constitute the first study of global temporal conditioning in sleep SSL, demonstrating that session-level context is a consistently useful signal across diverse frameworks and clinical tasks. Moreover, when a GTC-enabled variant achieves the best performance, the second-best result frequently comes from another GTC configuration rather than the unconditioned baseline, suggesting that the gains are robust across conditioning strategies rather than tied to a single placement.


\section{CONCLUSION}
We introduced GTC, a lightweight and architecture-agnostic mechanism that complements existing positional encodings by conditioning latent representations on the global time-of-sleep of each input window. Across three SSL frameworks, five clinically relevant tasks, and two independent pediatric cohorts, GTC generally improved F1 and AUPRC. These results highlight global temporal context as a broadly useful inductive bias for sleep representation learning. Future work could extend GTC to supervised settings and explore alternative session-level temporal signals.



\label{sec:results}

\section{Acknowledgement}
The Childhood Adenotonsillectomy Trial (CHAT)
was supported by the National Institutes of Health
(HL083075, HL083129, UL1-RR-024134, UL1 RR024989).
The National Sleep Research Resource was supported
by the National Heart, Lung, and Blood Institute (R24
HL114473, 75N92019R002). This work was partially supported by the AI Acceleration
Program at the University of North Carolina at Chapel
Hill.

\bibliographystyle{IEEEbib}
\bibliography{strings,refs}

\end{document}